\pdfoutput=1

\documentclass[11pt]{article}

\usepackage[table,dvipsnames]{xcolor}
\usepackage[]{acl2023}

\usepackage{times}
\usepackage{latexsym}
\usepackage{booktabs}
\usepackage{multicol}
\usepackage{tabularx}
\usepackage{graphics}
\usepackage{soul}
\usepackage{bbold}

\usepackage[T1]{fontenc}

\usepackage[utf8]{inputenc}

\usepackage{microtype}
\usepackage{paralist}
\usepackage{diagbox}
\usepackage{multirow}
\usepackage{comment}
\usepackage{booktabs}
\usepackage{relsize}
\usepackage{array}
\usepackage{multirow}
\usepackage{graphicx}
\usepackage{xspace}
\usepackage{cleveref}
\usepackage[position=top]{subfig}

\newcommand{\ex}[1]{\textit{#1}\xspace}

\newcolumntype{P}[1]{>{\centering\arraybackslash}p{#1}}

\title{Language models are not naysayers:\\An analysis of language models on negation benchmarks}

\author{Thinh Hung Truong$^{1}$ \qquad Timothy Baldwin$^{1,3}$ \qquad
\textbf{Karin Verspoor$^{2,1}$ \qquad Trevor Cohn$^{1,}$\Thanks{ Now at Google DeepMind.}   }\\
$^1$University of Melbourne\,\,\, $^2$RMIT University\,\,\, $^3$MBZUAI\\
\smaller \texttt{hungthinht@student.unimelb.edu.au} \,\,\,  \texttt{tb@ldwin.net} \\
\smaller \texttt{karin.verspoor@rmit.edu.au}\,\,\, \texttt{trevor.cohn@unimelb.edu.au}}

\begin{document}
\maketitle
\begin{abstract}

Negation has been shown to be a major bottleneck for masked language models, such as BERT.
However, whether this finding still holds for larger-sized auto-regressive language models (``LLMs'') has not been studied comprehensively.
With the ever-increasing volume of research and applications of LLMs, we take a step back to evaluate the ability of current-generation LLMs to handle negation, a fundamental linguistic phenomenon that is central to language understanding.
We evaluate different LLMs --- including the open-source GPT-neo, GPT-3, and InstructGPT --- against a wide range of negation benchmarks. 
Through systematic experimentation with varying model sizes and prompts, we show that LLMs have several limitations including insensitivity to the presence of negation, an inability to capture the lexical semantics of negation, and a failure to reason under negation. 

\end{abstract}

\section{Introduction}

Despite being a core linguistic phenomenon, negation remains a major stumbling block for modern NLP architectures \citep{kassner-schutze-2020-negated, hossain-etal-2022-analysis}.
A reason for this could be that texts containing negation are underrepresented in training data of language models, as humans tend to express themselves using affirmative rather than negative expressions \citep{ettinger-2020-bert}. 
Regardless, negation has been shown to be challenging even for humans to correctly interpret due to the diversity of forms across domains \citep{truong-etal-2022-improving}. 
For instance, in clinical documents, many acronyms are used to denote negation such as \ex{NAD 
(no abnormality detected)}, and implicit negation abounds, such as \ex{normal chest x-ray scan}, which implies the absence of an abnormality. 
Even more complex is the use of negation in combination with other linguistic phenomena such as quantifiers, gradable adjectives (\ex{not unattractive} does not imply \ex{attractive}) \citep{truong-etal-2022-another}; licensing context (negative polarity items, e.g.\ \ex{any, either, yet}, normally appear in certain negative grammatical contexts \citet{warstadt-etal-2019-investigating}); downward entailment (\ex{A man owns a dog} entails \ex{A man owns an animal} but \ex{A man does not own a dog} does not entail \ex{A man does not own an animal}) \citep{geiger-etal-2020-neural}.

Traditionally, negation has been treated as a standalone problem, e.g.\ as negation detection \citep{chapman2001simple}. The investigation of the impact of negation in various downstream tasks \citep{hossain-etal-2022-analysis, hossain2022leveraging}, or through probing \citep{ettinger-2020-bert} has revealed several limitations of modern large language models (``LLMs'') in handling negation.
Given that LLMs are being adopted in an ever-growing range of tasks and have been shown to display emergent abilities for high-level tasks that require complex reasoning  \citep{weiemergent}, we are interested in exploring how the handling of negation has progressed.

In this work, we investigate the performance of auto-regressive language models on different negation-focused benchmarks. 
Instead of just looking at samples containing negation in common NLP datasets, we consider datasets in which negation plays an important role in making the correct judgement.
In particular, we classify the benchmarks into three categories corresponding to the requisite negation reasoning abilities: (1) sensitivity to negation through cloze completion (fill-in-the-blank) queries of factual statements; (2) lexical semantics of negation through classification of antonym/synonym relationships; and (3) ability to reason with negation through language inference tasks.

\begin{table*}[!t]
\footnotesize
    \centering
    \begin{tabular}{p{1.5cm} p{2cm} p{2cm} p{8cm} }
    \toprule
      \bf Benchmark  & \bf Task & \bf \# Samples & \bf Example\\
      \midrule
      MKR-NQ  & Completion & 3360 & Query: \ex{Iburofen isn't a kind of [MASK].} Wrong completions: \ex{NSAID, painkiller, drug, medicine.} \\
      \midrule
      MWR  & Completion & 27546 & Query: \ex{Demand is an antonym of [MASK].} Wrong completions: \ex{necessitate, demands, request, requirement, imposition, need, demand.}\\
      \midrule
      SAR &  NLI & 2000 & Word 1: \ex{Superiority} / Word 2: \ex{Inferiority} / Label: Antonym \\
      \midrule
      NegNLI  & NLI & 4500 & P: \ex{They watched me constantly for weeks.} / H: \ex{They did not leave me on my own for weeks.} / Label: Entailment \\
      \midrule
      NaN-NLI  & NLI & 258 & P: \ex{Not all people have had the opportunities you have had.} / H: \ex{Some people have not had the opportunities you have had.} / Label: Entailment  \\
      \midrule
      MoNLI & NLI & 200 & P: \ex{The man does not own a dog.} / H: \ex{The man does not own a mammal.} / Label: Not Entailment  \\
      \bottomrule
    \end{tabular}
    \caption{Summary of the negation-related benchmark datasets used in this paper.}
    \label{tab:benchmark}
\end{table*}

We conduct extensive experiments using prompt-based learning to facilitate zero- and few-shot evaluation of LLMs, and find the following:
\begin{itemize}
    \item larger LMs are more insensitive to negation compared to smaller ones (\Cref{sec:res-presence-negation});
    \item LLMs lack lexical semantic knowledge about negation, yielding almost random performance for synonym/antonym classification (\Cref{sec:res-lexical-semantic});
    \item LLMs have limited ability to reason under negation, performing worse than random across most NLI datasets (\Cref{sec:results-nli}). Only with the latest instruction fine-tuned model \citep{ouyang2022training, chung2022scaling} do we observe above-chance performance (\Cref{sec:res-gpt3});
    \item For each dataset, we also experiment with prompt variations and find that in most cases, providing more information (context, instruction, simple wording) leads to a degradation in performance.
\end{itemize}


\section{Experimental settings}

In this section, we outline the settings that , including benchmark datasets, models to evaluate, and the prompts that were used. Our code is available at \url{https://github.com/joey234/llm-neg-bench}.

\begin{table*}[!t]
\footnotesize
    \centering
    \begin{tabular}{p{1cm} p{1.5cm} p{11cm}}
    \toprule
      \bf Task  & \bf Prompt name & \bf Example \\
      \midrule
      MKR-NQ & Default & An expectorant isn't a type of \underline{\hspace{0.5cm}}  \\
      \cmidrule(lr){2-3}
        & Contrasting & \textbf{An expectorant is a type of medicine.} An expectorant isn't a type of \underline{\hspace{0.5cm}} \\
      \cmidrule(lr){2-3}
      & Discourse & \textbf{An expectorant is a type of medicine. Therefore,} an expectorant isn't a type of \underline{\hspace{0.5cm}} \\
      \cmidrule(lr){2-3}
      
        & Mask & An \textbf{[MASK]} is a type of medicine. An \textbf{[MASK]} isn't a type of \underline{\hspace{0.5cm}} \\
        \midrule
      MWR & Default  & Greed is an antonym of \underline{\hspace{0.5cm}} \\
      \cmidrule(lr){2-3}
          
          
          & Quote  & \textbf{The word ``}greed\textbf{''} is an antonym of \textbf{the word `` \underline{\hspace{0.5cm}}} \\
          \midrule
    SAR & Default & Choose the correct answer: bad and good are antonyms or synonyms? Answer: \underline{\hspace{0.5cm}} \\
      \cmidrule(lr){2-3}
        
        & Simple & \textbf{Choose the correct answer:} bad and good are \textbf{opposite} or \textbf{similar}? \textbf{Answer:} \underline{\hspace{0.5cm}} \\
      \cmidrule(lr){2-3}
        
        & Negation & \textbf{Antonyms are words with opposite meaning. Synonyms are words with similar meaning. Choose the correct answer:} bad and good are antonyms or synonyms? \textbf{Answer:} \underline{\hspace{0.5cm}} \\ 
        
        \midrule
    NLI & Default & Not all people have had the opportunities you have had. \\
        &  & \textbf{Question:} Some people have not had the opportunities you have had. True, False, or Neither? \\
        & & \textbf{Answer:} \underline{\hspace{0.5cm}} \\
      \cmidrule(lr){2-3}
        
        & Negation & \textbf{The question requires reasoning about negation.} \\
        &  & Not all people have had the opportunities you have had. \\
        &  & \textbf{Question:} Some people have not had the opportunities you have had. True, False, or Neither? \\
        & & \textbf{Answer:} \underline{\hspace{0.5cm}} \\
      \bottomrule
    \end{tabular}

    \caption{Prompts used for each task}
    \label{tab:prompts}
\end{table*}

\subsection{Benchmarks}

We use a range of benchmark datasets that exhibit the effects of negation across a wide range of tasks, in the form of either cloze completion or classification tasks. An overview of the datasets is presented in \Cref{tab:benchmark}, categorized according to purpose and  the type of negation they contain. Specifically, we focus on: (1) investigating whether LLMs are sensitive to the presence of negation in factual statements; (2) testing whether LLMs capture negation in lexical semantics relations (synonym/antonym relations); and (3) investigating whether LLMs are able to reason under negation through multiple natural language inference benchmarks. We discuss the datasets in greater detail in \Cref{sec:finding}.

\subsection{Models}

For the LLMs, we primarily focus on open-source auto-regressive LLMs with up to 6.7B parameters, including GPT-Neo \cite{gpt-neo}, and OPT \citep{zhang2022opt}, which are claimed to be comparable in performance to similar-sized GPT-3 class models. 
Architecture-wise, they are both decoder-only PLMs pre-trained with a causal LM objective, with the main difference being in their pre-training corpora: GPT-neo was trained solely on the Pile dataset \citep{gao2020pile} consisting of 22 sub-datasets spanning different sources, whereas OPT was trained on the combination of datasets used in RoBERTa \citep{liu2019roberta}, Pile, and PushShift Reddit \citep{baumgartner2020pushshift}.
We use the official  model checkpoints from HuggingFace hub,\footnote{\url{https://huggingface.co/models}} as detailed in \Cref{app:model-checkpoints}.
We experiment with smaller-sized variants of these two classes of models to observe the effect of scaling on their performance over different benchmarks.

We also consider base GPT-3 (175B) \citep{brown2020language}, and its instruction fine-tuned variant InstructGPT \citep{ouyang2022training}, as well as a strong  open-source instruction-tuned model FLAN-T5-XXL (11B) \citep{chung2022scaling}, to examine how recent commercial LLMs perform on negation.  

\subsection{Prompts}

We adopt prompt-based learning to enable zero- and few-shot evaluation of LLMs \citep{radford2019language}.
Given that LLMs have been found to be sensitive to prompt variation \citep{wei2021finetuned}, and that more natural-sounding prompts 
correlate with model performance \citep{gonen2022demystifying}, we also experiment with different types of prompts (see \Cref{tab:prompts}).

We use GPT-3 style prompts \citep{brown2020language} as the \textit{Default} setting. 
As handling negation plays an important role in all tasks, we also design prompts to prime the LLMs to focus more on the negation context, by introducing modifications specific to each task.
In detail, for the cloze completion task MKR-NQ, we investigate whether a given model can detect the difference between two contrasting sentences (with/without negation). To achieve this, we prepend the prompt with the corresponding sentence without negation (\textit{Contrasting} prompt).
In addition, we also evaluate alternative prompts where we connect the two sentences with a discourse marker (\textit{Discourse} prompt), or mask the main subject to encourage the model to attend more to negation cues (\textit{Mask} prompt).

For antonym/synonym-related tasks (MWR, SAR), we also experiment with simplifying the prompt and use descriptive terms rather than the formal names of the relations (e.g.\ \ex{antonyms, synonyms} $\rightarrow$ \ex{opposite of, similar to}), based on the intuition that these terms will appear more frequently in the pre-training data. 

Finally, for classification tasks, we propose negation-aware prompting (\textit{Negation} prompt) by modifying the prompts to explicitly state that the task involves reasoning about negation. 
Note that we explicitly include class options in the prompts to help reduce the effect of the surface form competition causing LLMs to assign lower probabilities to the correct answers \citep{holtzman-etal-2021-surface}.

For datasets with an accompanying training set (SAR, MoNLI), we also experiment with few-shot evaluation formulated as \textit{in-context learning} by prepending the input prompts with 10 random samples from the training set.

\subsection{Metrics}

To evaluate cloze completion tasks, we employ \textit{Weighted Hit Rate (WHR)} \citep{jang-etal-2022-beyond}, which measures the number of matches between the top-k predicted words and a given set of target wrong predictions, taking into account the prediction probabilities:
\begin{equation}
    \textit{WHR}_{k}{(x, W)} = \frac{\sum^{k}_{i=1}{c_i \times \mathbb{1}(w_i \in W_x)}}{\sum^{k}_{i=1}c_i}
\end{equation}
where $W_x$ is the wrong prediction set of the input query $x$, and $w_i$ is the top $i$-th prediction with confidence score $c_i$, obtained by taking the softmax of log probabilities $p(w_i|x)$ from the LM. Note that the model performance is better if there are fewer matches between models' predictions and wrong completions, \textit{WHR} is an error metric (lower is better).
One problem with the \textit{WHR} metric is that we can only evaluate using a fixed set of wrong predictions. Regardless, we believe the relative performance numbers are indicative of model performance. 

For classification tasks, we evaluate using \textit{Accuracy}, noting that all datasets are reasonably balanced.

\section{Main findings}
\label{sec:finding}
We summarize the main findings in this section. 
In general, the performance of GPT-neo and OPT follows a similar trend across all benchmarks (we present GPT-neo results; results of OPT models are in \Cref{app:opt}).

\subsection*{Finding 1: Larger LMs are more insensitive to negation}
\label{sec:res-presence-negation}


\paragraph{MKR-NQ \citep{jang-etal-2022-beyond}} Masked Knowledge Retrieval -- Negated Query (MKR-NQ)  is a negated version of the LAMA dataset \citep{petroni-etal-2019-language}, which contains lexicalized statements of triples in ConceptNet \citep{speer2017conceptnet}. This dataset contains factual statements with verbal negations (i.e.\ negators \ex{not, don't} are associated with the main verb of the sentence), e.g.\ \ex{Iburofen is a type of medicine. $\rightarrow$ Iburofen isn't a type of medicine}. 

Each sample contains the query along with a set of wrong word completions, supporting the evaluation of the sensitivity of the model to negation by measuring how likely a model will generate incorrect completions.
Note that MKR-NQ only considers sample sentences that contain a single verb, making it trivial to negate the original sentences.

\begin{figure*}
\centering
    \includegraphics[width=\linewidth]{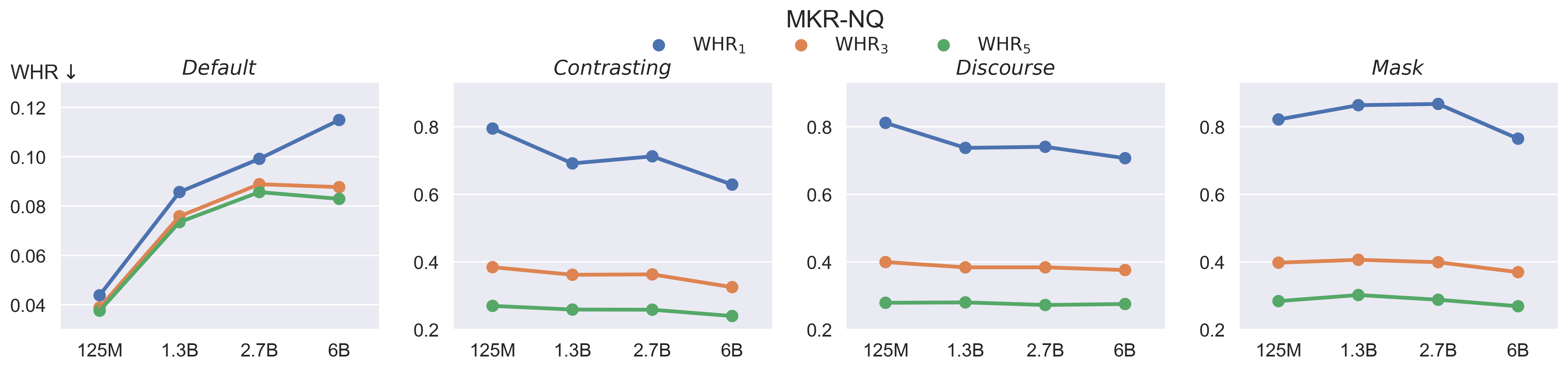} 
    \caption{Zero-shot performance of GPT-neo on MKR-NQ using different prompts under the Weighted Hit Rate (WHR) metrics (lower scores are better). Note the different scale for the left-most plot.}
    \label{fig:ele-mkr-prompts}
\end{figure*}

\paragraph{Findings} From \Cref{fig:ele-mkr-prompts}, which is based on LLMs with a negated factual statement (\textit{Default} prompt), we observe relatively low hit rates ($<$ 0.15) across all model sizes, and a clear inverse scaling trend between model sizes and their performance.
The smallest variant (\texttt{GPT-neo-125M}) has the best performance, which is comparable to that of masked language model of a similar size (\texttt{BERT-base}, 110M parameters) \citep{jang-etal-2022-beyond}.
This phenomenon reflects the finding that larger models tend to memorize the training data more \citep{inverse-scaling-prize,jang2022can}.
Moreover, higher hit rates for top-1 predictions suggest that models predict wrongly with high confidence.

For \textit{Contrasting} prompts, in which we prepend the negated statement with its non-negated version, we notice a drastic increase in \textit{WHR}, showing that models are prone to repeating what is presented in the prior context, confirming the finding of \citet{kassner-schutze-2020-negated}.
When a discourse term is added to connect the two sentences (\textit{Discourse} prompt), we do not observe any improvement, and the performance of the largest model is even worse.
To investigate whether this phenomenon is attributable to models not being able to detect the presence/absence of negation, we experiment with masking out the main noun/verb of the queries (\textit{Mask} prompt). 
We observed even higher \textit{WHR}, especially for the top-1 prediction in this setting.
The results suggest that repetitions are caused more by LLMs being easily primed by repeating what is present in the previous context, than by generating words that are closely associated with the main subject of interest. This again shows that the models cannot differentiate between identical contexts, differing only on whether negation is present or absent (i.e., outputs tend to be similar with or without negation).

To further analyze the outputs, we calculate the perplexity (PPL) of the generated predictions to determine their plausibility \citep{wilcox-etal:2020-on-the-predictive-power}.
Here, we choose the model with the best $\textit{WHR}_5$ score on the MKR-NQ benchmark, and calculate the mean perplexity over all queries for each prompt type (5 completions for each query).
PPL is calculated as the exponentiated average negative log-likelihood of a sequence, with exponent base $e$.
As a point of reference, we calculated the average perplexity of the provided completion of the original non-negated dataset (denoted \textit{Corpus}).
From the reported perplexities (\Cref{tab:ppl}), we can see that \textit{Default} output are the most plausible (with PPL markedly lower than \textit{Corpus}), while \textit{Contrasting} is the least natural.
The remaining prompts types (\textit{Discourse, Mask}) are comparable to \textit{Corpus}.
These results show that LLMs can indeed generate plausible and human-like output for this task.

\begin{table}[!t]
\footnotesize
    \centering
    \begin{tabular}{p{1.25cm} p{4cm} p{0.75cm}}
    \toprule    
    Setting & Example & Mean PPL$\downarrow$  \\
    \midrule
        Corpus & [Baseball is a type of sport.] &  434.42 \\
        \midrule
        \textit{Default} & [Baseball isn't a type of sport.] & 288.94 \\
        \textit{Contrasting} & Baseball is a type of sport. [Baseball isn't a type of sport.] & 533.56 \\
        \textit{Discourse} & Baseball is a type of sport. Therefore, [baseball isn't a type of sport.]  & 477.44  \\
        \textit{Mask} & MASK is a type of sport. [MASK isn't a type of sport.] & 448.23  \\
    \bottomrule
    \end{tabular}
    \caption{Mean perplexity (PPL) calculated using the \texttt{GPT-J-6B} model. Only the strings enclosed in square brackets are considered during calculation in order to provide a fair comparison with similar token length. For Corpus, PPL is calculated using the provided gold completion.}
    \label{tab:ppl}
\end{table}

\subsection*{Finding 2: LMs fail to capture synonym/antonym lexical relations}
\label{sec:res-lexical-semantic}

\paragraph{MWR \citep{jang-etal-2022-beyond}}  To test the ability of LMs to capture negative lexical semantics, we use MWR dataset, where models are asked to predict the antonym/synonym of a target word. The dataset was constructed by using the most frequent nouns, adjectives, and adverbs that appear in SNLI \citep{bowman-etal-2015-large}, then choosing their corresponding synonyms and antonyms from ConceptNet \citep{speer2017conceptnet}.
The dataset also contains different wordings for antonym-asking and synonym-asking queries (e.g.\ \ex{is the opposite of, is different from} and \ex{is similar to, is a rephrasing of}) to test model sensitivity to prompt variations.

\begin{figure}[!t]
\centering
    \includegraphics[width=\columnwidth]{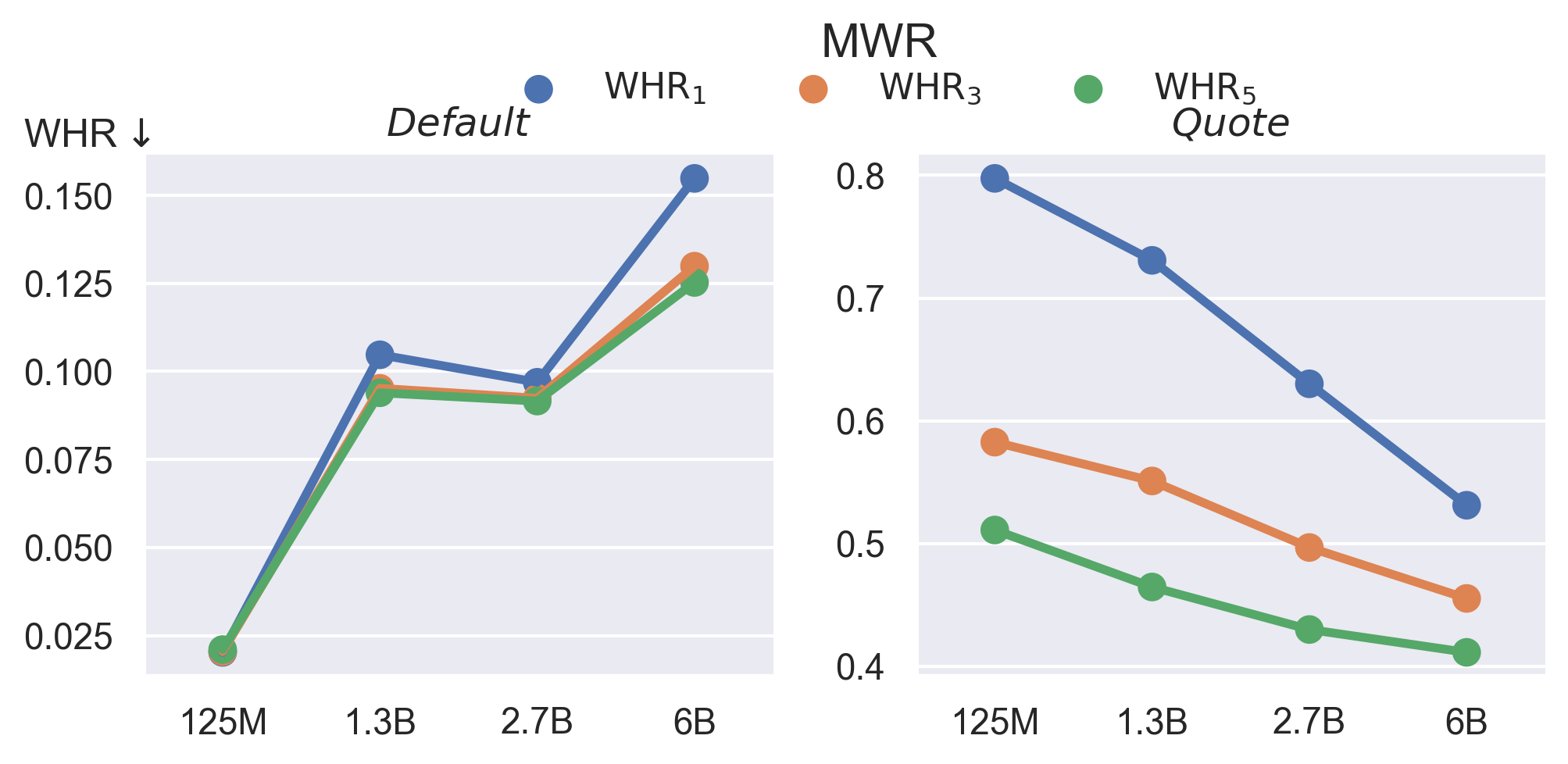} 
    \caption{Zero-shot performance of GPT-neo on MWR using different prompts (WHR metrics; lower is better)}
    \label{fig:ele-mwr-prompts}
\end{figure}

\paragraph{Findings} From \Cref{fig:ele-mwr-prompts}, we can observe the same inverse scaling trend as for MKR-NQ using the \textit{Default} prompt, where the hit rate of the smallest model is around 0.02, better than previously-reported SOTA results \citep{jang-etal-2022-beyond}.
With a more natural query with more focus on the target words via quotation marks (\textit{Quote} prompt), surprisingly, we noticed a drastic jump in hit rates.
However, MWR may not be a good indicator of model performance due to how the task is framed.
One problem is that models can generate words that are not in the given wrong prediction set, but are also irrelevant, and are also neither antonyms nor synonyms of the given target words, as demonstrated in \Cref{tab:mwr_output}.

\begin{table}[!t]
\footnotesize
    \centering
    \begin{tabular}{p{2cm} p{2cm} p{2cm}}
    \toprule    
    Query & Wrong completions & Top-5 predictions  \\
    \midrule
    Greed is an antonym of & \ex{greed, avarice, desire, greeds, gluttony} & \ex{altruism, self-sacrifice, self-denial, self-abnegation, \textbf{gods}} \\
    Finale is an antonym of & \ex{conclusion, finish, finales, finale} & \ex{\underline{last}, \textbf{epiphany}, \underline{finality}, \textbf{anti-climax}, \textbf{anti-climactic}} \\
    
    \bottomrule
    \end{tabular}
    \caption{Example output of \texttt{GPT-J-6B} on MWR. \textbf{bolded} words are related to target words, but are neither synonyms nor antonyms. \underline{underlined} are wrong antonyms but are not in the given set of wrong completions.}
    \label{tab:mwr_output}
\end{table}

\begin{figure}[!t]
    \centering
    \includegraphics[width=\columnwidth]{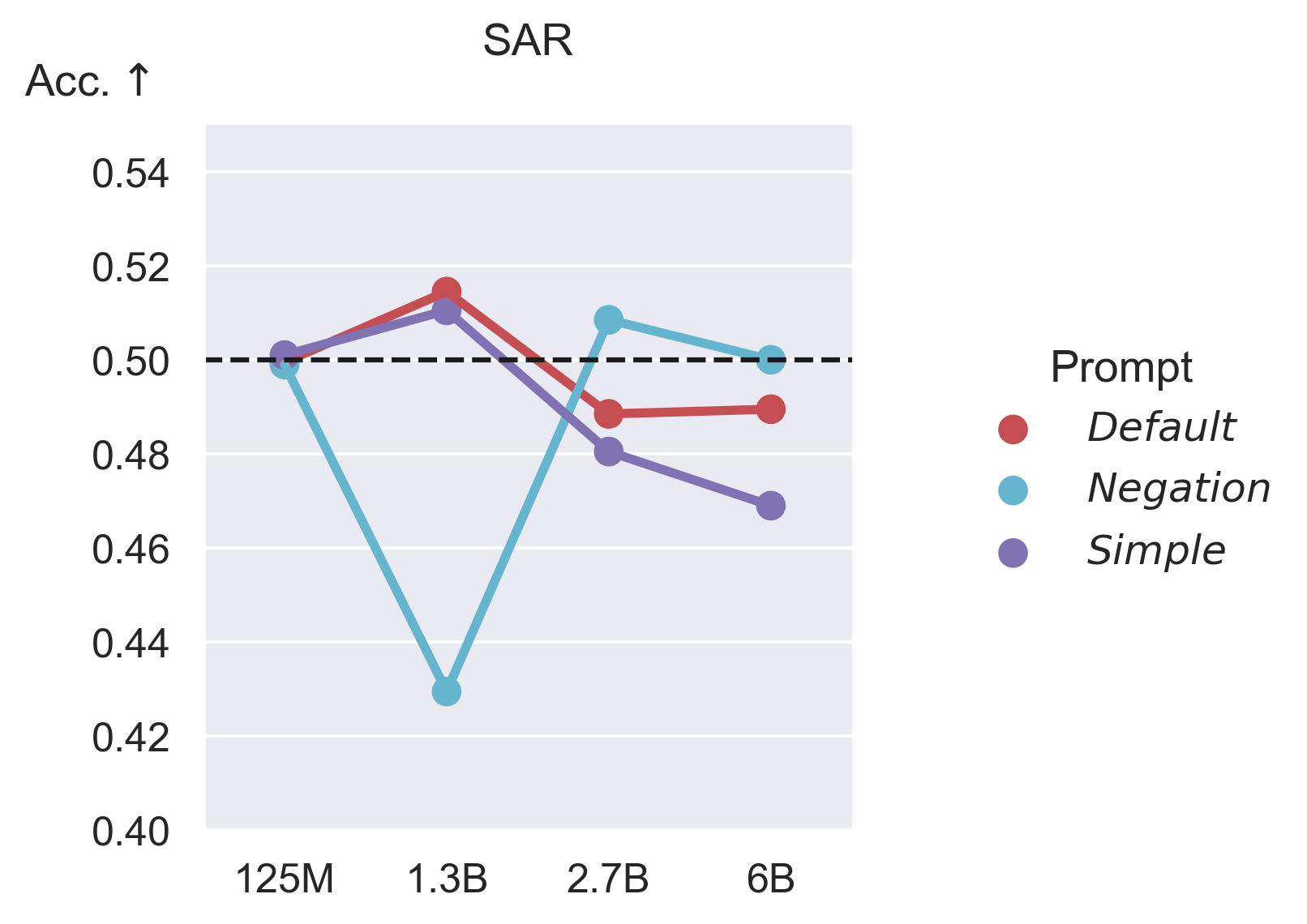}

    \caption{Zero-shot performance of GPT-neo on SAR dataset using different prompts (accuracy metric; higher is better)}
    \label{fig:ele-sar}
\end{figure}

\paragraph{SAR \citep{jang-etal-2022-beyond}} To further investigate the ability of LLMs to capture negative lexical semantics, we consider the antonym/synonym relation classification task (SAR). Different from the MWR cloze-style synonym/antonym prediction task, this benchmark is framed as a binary classification task 
of predicting 
the correct antonym or synonym relationship between two given words.
Data is once again taken from ConceptNet, where triplets with synonym and antonym relations are extracted in equal numbers (1000 samples for each relation).

\paragraph{Findings} In contrast to the high results for MWR, we find that for this task, model performance is equivalent to random, with accuracy fluctuating around 0.5 (\Cref{fig:ele-sar}).
For prompt variants, we do not observe any meaningful improvement, in that \textit{Simple} follows a similar trend to \textit{Default} and \textit{Negation} performs better for larger models (2.7B and 6B).
This is a huge degradation from previous fully fine-tuned results over encoder models.
For instance, \citet{jang-etal-2022-beyond} reported that BERT\textsubscript{large} achieves 92.5\% accuracy on SAR.
We argue that this is a specific task that is not captured in the next token prediction training objective of LLMs and thus, requires explicit supervision.

\begin{figure*}[!t]
    \centering
    \includegraphics[width=0.8\linewidth]{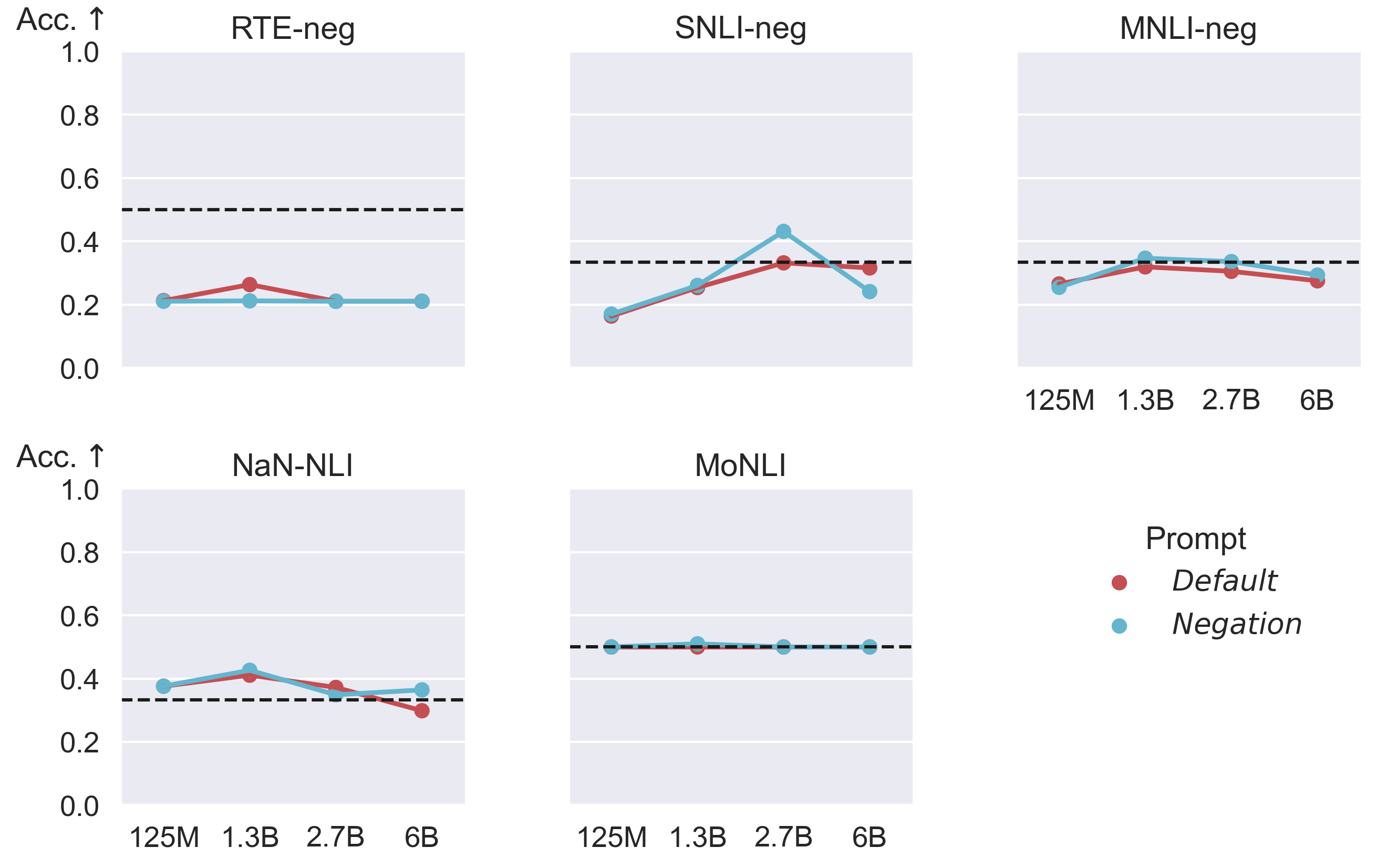}%

    \caption{Zero-shot performance of GPT-neo on NLI datsets using different prompts (higher is better). The dashed line denotes a random baseline. Note that RTE-neg and MoNLI are 2-way classification tasks while the rest are 3-way.}
    \label{fig:ele-nli}
\end{figure*}

\begin{table*}[!t]
\footnotesize
    \centering
    \begin{tabular}{p{1.4cm} P{0.2cm} P{1.5cm} P{2cm} P{2cm} P{2cm} P{2.2cm} }
    \toprule
        Benchmark & & GPT-J-6B & GPT-3  & InstructGPT & InstructGPT w/ \textit{Neg.} prompt & FLAN-T5-XXL w/ \textit{Neg.} prompt \\
        \midrule
        MKR-NQ & \multirow[b]{1}{*}{\rotatebox[origin=c]{90}{ \scriptsize \textit{WHR}\textsubscript{5} }$\downarrow $ }   & \textbf{0.083} & 0.172 & 0.195 & NA & NA   \\         
        MWR & & \textbf{0.125} & 0.488 & 0.504 & NA & NA  \\
        \midrule
        SAR & \multirow{6}{*}{\rotatebox[origin=c]{90}{\footnotesize \textit{Accuracy}}$\uparrow$} & 0.490 & 0.501 &  0.687 & \textbf{0.780} & 0.507  \\
        SNLI-neg & & 0.316 & 0.267  & 0.640 & \textbf{0.673} & 0.477 \\ 
        MNLI-neg & & 0.275 & 0.359  & 0.548 & \textbf{0.625} & 0.354 \\
        RTE-neg & & 0.211 & 0.525  & 0.767 & \textbf{0.807} & 0.770 \\ 
        NaN-NLI & & 0.298 & 0.469  & 0.647 & \textbf{0.682} & 0.376 \\
        MoNLI & & 0.500 & \textbf{0.540} & 0.470 & 0.400 & 0.500 \\
        \bottomrule
        
    \end{tabular}
    \caption{Zero-shot results on the different benchmarks. ``NA'' denotes that \textit{Negation} prompts are not applicable to MKR-NQ and MWR. The best results are bolded for each task (row).}
    \label{tab:gpt3-results}
\end{table*}

\subsection*{Finding 3: LLMs are unable to reason under negation}
\label{sec:results-nli}

\paragraph{NegNLI \citep{hossain-etal-2020-analysis}}  NegNLI contains 4500 premise--hypothesis pairs with \textit{important} negation, where negation is essential in making the correct judgement about the relationship between the premise--hypothesis pairs. 
Samples are extracted from the commonly-used NLI datasets (RTE \citet{dagan2005pascal}, SNLI \citet{bowman-etal-2015-large}, MNLI \citet{williams-etal-2018-broad}), then the negator \ex{not} is added to the main verb either in the premise, hypothesis, or both. Here, we consider each subset separately, as the number of classes are not the same, and denote them SNLI-neg, MNLI-neg, RTE-neg.
 
\paragraph{MoNLI \citep{geiger-etal-2020-neural}} MoNLI  is an NLI dataset focused on lexical entailment and negation. Specifically, the dataset investigates the downward monotonicity property where negation reverses entailment relations (e.g.\ \ex{dance} entails \ex{move}, but \ex{not move} entails \ex{not dance}). MoNLI was created by extending samples from SNLI by substituting the nouns by their hypernyms/hyponyms from WordNet \citep{miller1998wordnet}.

\paragraph{NaN-NLI \citep{truong-etal-2022-another}} NaN-NLI  is a test suite which focuses on sub-clausal negation, in which only part of the sentence's meaning is negated, thus making it harder to correctly determine the correct negation scope (e.g.\ in \ex{Not the first time that they pulled that off} the negation scope is only \ex{Not the first time} and the main clause of the sentence \ex{they pulled that off} is not negated).
Each premise--hypothesis pair is constructed so that the corresponding hypotheses are constructed to reflect different interpretations that the negated instance in the premise are likely to be misunderstood for.

\begin{figure}[!t]
    \centering
    \includegraphics[width=\columnwidth]{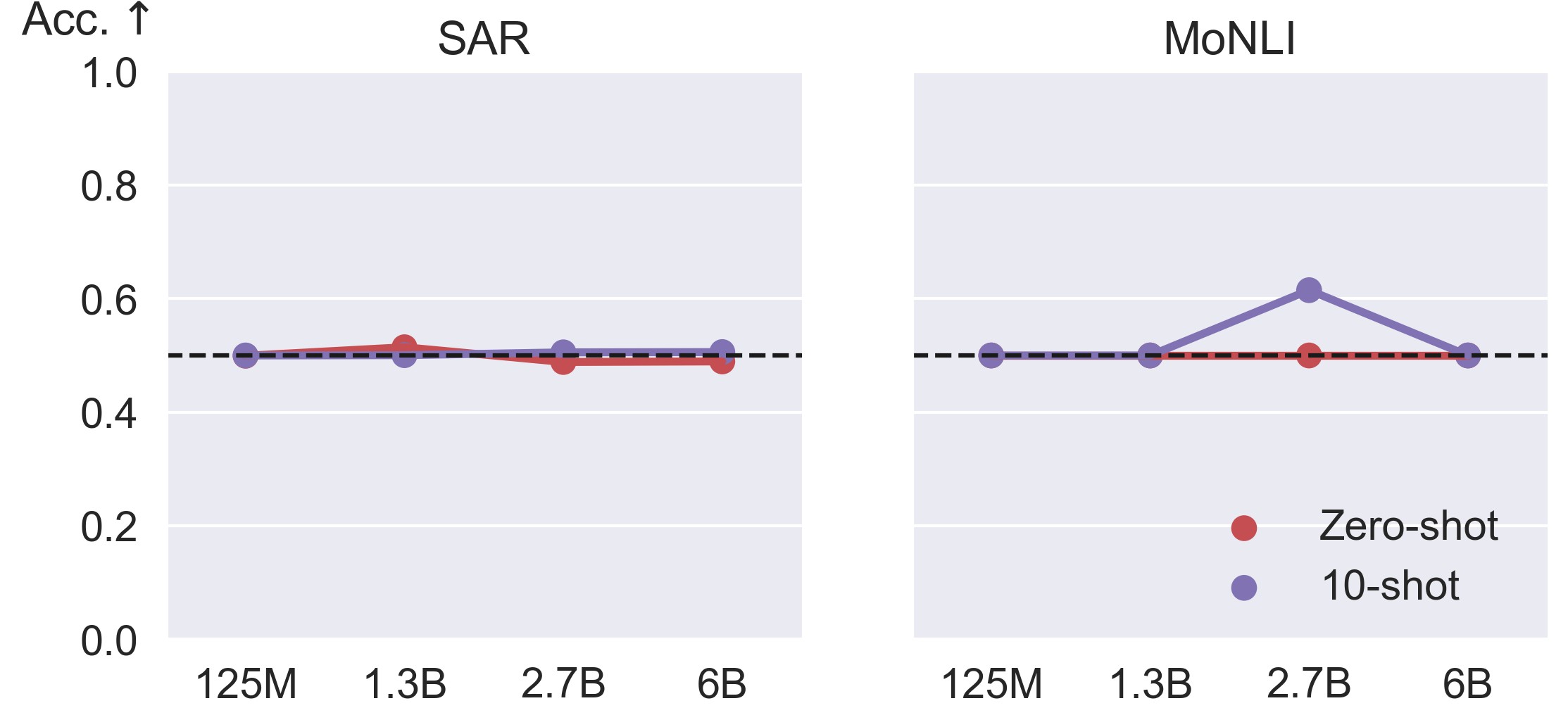}%

    \caption{10-shot performance of GPT-neo on SAR and MoNLI using \textit{Default} prompt (higher is better)}
    \label{fig:ele-fewshot}
\end{figure}

\paragraph{Findings} Similar to the antonym/synonym classification task, the performance for most negation-focused NLI benchmarks is low.
In particular, for all NLI datasets, the performance is generally lower than baseline.
As shown in \Cref{fig:ele-nli}, scaling up model size has almost no effect, and the largest model performs worse in many cases, even when the prompt explicitly states that the task requires reasoning about negation (\textit{Negation} prompt).
For datasets which include a training set (SAR, MoNLI), we also experimented with few-shot learning but did not observe any noticable improvement (\Cref{fig:ele-fewshot}). One exception is that the 2.7B model seems to pick up some signal from the provided MoNLI training samples, but falls back again when we increase the model size to 6B.
 
Even with general NLI datasets, zero-shot applications of LLMs were previously shown to be roughly equivalent to a random baseline \citep{wei2021finetuned}. 
When negation is involved, the task becomes even more complex.
As pointed out in \citet{brown2020language}, one possible reason that LLMs struggle with NLI is that the samples consist of two disjoint  sentences, which are unlikely to appear naturally in standard training corpora. 
We hypothesise that NLI is a generally hard task that requires substantially more supervision in order for models to detect meaningful patterns.

\subsection*{Finding 4: Instruction fine-tuning improves reasoning under negation}
\label{sec:res-gpt3}

We further evaluate with GPT-3 class models of significantly larger scale (175B), which have been shown to achieve strong results in zero- and few-shot settings across a wide range of tasks \citep{brown2020language}.
In detail, we benchmark the largest GPT-3 model (\texttt{text-davinci-001}: \citet{brown2020language}) and its variant InstructGPT, which is trained to follow human instructions using reinforcement learning (\texttt{text-davinci-003}: \citet{ouyang2022training}).
The results can be found in \Cref{tab:gpt3-results}. 

For the base GPT-3 model, the results over most benchmarks are no better than much smaller language models (\texttt{GPT-neo-125M}).
For cloze-completion tasks, consistent with the earlier-observed trend of larger models performing worse, we observe higher (worse) \textit{WHR} scores compared to that of smaller language models, confirming our finding that larger models are more \textit{in}sensitive to the presence of negation. Results get even worse with using the instruction fine-tuned model. 

\begin{figure}[!t]
    \centering
    \includegraphics[width=\columnwidth]{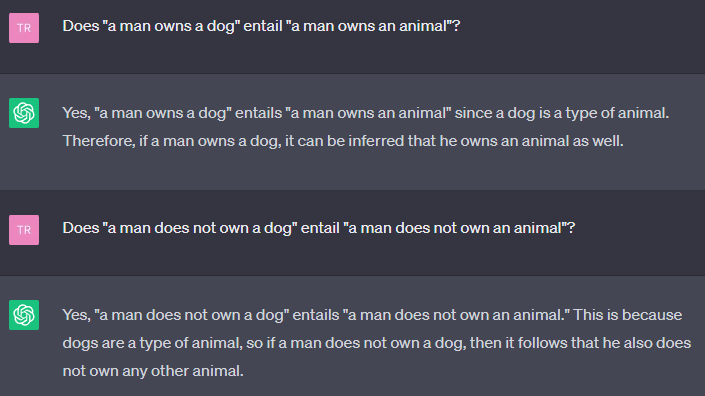}

    \caption{A ChatGPT-generated output of a failed negative monotonicity reasoning sample. The output was generated using ChatGPT Feb 13 Version}
    \label{fig:chatgpt-monotonicity}
\end{figure}

On the other hand, for most classification tasks, InstructGPT achieves better zero-shot results than other models.
In addition, using this model in combination with explicit instruction about negation (\textit{Negation} prompt) further improves performance, which we did not observe for other LLMs.
It is, however, unclear what data the instruction-tuning process was performed on. Thus, the huge gain in performance could be attributed to the existence of similar patterns in the training set (i.e.\ explicit supervision over similar tasks). Interestingly, InstructGPT performance on MoNLI did not increase (it underperfomed other models). We hypothesize that this is due to an inductive bias from model's ability to reason with hypernymy. For instance, the model can understand that ``\textit{dog} is an \textit{animal}'' (and therefore \textit{own an animal} entails \textit{own a dog}), but incorrectly generalizes this logic to a similar sample containing negation (\textit{not own a dog} entails \textit{not own an animal}). This is indeed true when we look at the explanation generated by ChatGPT, the subsequent model to InstructGPT (\Cref{fig:chatgpt-monotonicity}).

We also experiment with the instruction-tuned FLAN-T5-XXL model \citep{chung2022scaling} and find that the results are better than GPT-3 for most NLI tasks, despite being $\sim$16x smaller.
These results suggest that instruction fine-tuning has much greater impact than model scaling in terms of models developing the ability to perform reasoning tasks under negation.

\section{Related work}
\label{sec:related_work}

Our work builds upon previous research on negation. In particular, we were inspired by the pioneering works of \citet{kassner-schutze-2020-negated} and \citet{ettinger-2020-bert}, which reveal that pre-trained language models have a major issue in being insensitive to the presence of negation, based on evaluation over a set of cloze-style queries.
Following this line of research, \citet{jang-etal-2022-beyond} also explored negation in a cloze completion context by negating factual statements extracted from ConceptNet and come to a similar finding.

In a broader context, \citet{hossain-etal-2020-analysis,hossain-etal-2022-analysis} investigated the performance of BERT-based methods on samples containing negation in the GLUE \citep{wang-etal-2018-glue} and SuperGLUE \citep{wang2019superglue} datasets. Their main finding is that the results for the subsets containing only negation are lower than those without, as well as the whole test set, showing that models struggle with negation, even when fine-tuned on relevant training data. 
\citet{ravichander2022condaqa} proposed the challenging CONDAQA dataset to test the ability of models to reason about the implications of negation. 
The authors conducted comprehensive analysis of different types of LLMs under different settings, and found that the best-performing models were still well below human performance.
Negation has also been investigated as part of psycholinguistic probing datasets \citep{lialin-etal-2022-life,jumelet-etal-2021-language,staliunaite-iacobacci-2020-compositional}.
Contrasting previous finding, \citet{gubelmann-handschuh-2022-context} found that the ability to understand negation of LMs is underestimated in previous studied.
Through designing a controlled dataset with minimal pairs varying in syntactic structure, gender, profession, and first name, they concluded that the models are indeed sensitive to negation and thus, their struggle comes more from the contextualization of the tasks.

As part of the analysis on emergent abilities of LMs, negation has been shown to be one of the tasks that displays a flat scaling curve \citep{weiemergent} or even inverse-scaling \citep{inverse-scaling-prize}. This behaviour was later shown to be alleviated by instruction fine-tuning \citep{wei2022inverse}.
The effectiveness of instruction fine-tuning is further supported in \citet{jang2023consistency}. 
The authors investigated the logical consistency of ChatGPT and found that ChatGPT understands negation and antonyms much better than previous models.

Beside probing and evaluation, there have also been works on making language models more robust to negation, including unlikelihood training \citep{hosseini-etal-2021-understanding}, adaptive pre-training on relevant data \citep{truong-etal-2022-improving}, leveraging affirmative interpretations from negation \cite{hossain-blanco-2022-leveraging}, and learning better representation of negation through contrastive learning \citep{jiang-etal-2022-promptbert, wang2022sncse}.

\section{Conclusion}

We have shown that LLMs still struggle with different negation benchmarks through zero- and few-shot evaluations, implying that negation is not properly captured through the current pre-training objectives.
With the promising results from instruction-tuning, we can see that rather than just scaling up model size, new training paradigms are essential to achieve better linguistic competency.
Through this investigation, we also encourage the research community to focus more on investigating other fundamental language phenomena, such as quantification, hedging, lexical relations, and downward entailment.

\section{Limitations}

First, regarding the experimental settings, the \textit{WHR} metrics used to evaluate cloze completion tasks are imperfect, as we discussed. Framing cloze completion tasks in the style of multiple-choice question answering to limit the options that models are evaluated on would be a good direction to follow \citep{robinson2022leveraging}. In addition, the prompt engineering in this work is in no way exhaustive, and could be extended using different prompt engineering strategies such as soft prompt tuning \citep{lester-etal-2021-power}, or mining- and paraphrasing-based methods to generate high quality prompts \citep{jiang-etal-2020-know}.

Second, due to computational constraints, we could not perform an extensive set of experiments for larger models like PaLM (with up to 540B parameters) \citep{chowdhery2022palm}. Recent work by \citet{wei2022inverse} has shown that the inverse scaling trend on several benchmarks can be alleviated using the large instruction fine-tuned models such as FLAN-PaLM-540B, which is largely in line with our findings regarding InstructGPT and FLAN-T5. With a small-scale experiment, we found that ChatGPT displayed strong performance on challenging samples in the investigated benchmark, so the main findings of the paper may not hold true for newer LLMs.

Finally, this work only considers negation in the English language. There is every reason to believe that negation is an equally challenging problem in other languages. As this is a linguistically-intensive task, and requires native speakers to conduct thorough analysis of the results, we leave this for future work.

\section*{Acknowledgement}

The authors would like to thank the anonymous reviewers for their detailed, kind, and constructive reviews.
This research was conducted by the Australian Research Council Training Centre in Cognitive Computing for Medical Technologies (project number ICI70200030) and funded by the Australian Government.
This research was undertaken using the LIEF HPC-GPGPU Facility hosted at the University of Melbourne. This Facility was established with the assistance of LIEF Grant LE170100200.

\bibliography{anthology,custom}
\bibliographystyle{acl_natbib}


\appendix

\section{Model checkpoints}
\label{app:model-checkpoints}
For open-sourced LMs, we consider the official released checkpoints on the HuggingFace hub at:
\begin{itemize}
\footnotesize
    \item \url{https://huggingface.co/EleutherAI/x}
    
    \item \url{https://huggingface.co/facebook/y}
\end{itemize} 
where \texttt{x} in \{\textit{gpt-neo-125M,gpt-neo-1.3B,gpt-neo-2.7B,gpt-j-6B}\}, and \texttt{y} in 
\{\textit{opt-125m,opt-350m,opt-1.3b,opt-2.7b,opt-6.7b}\}.

For GPT-3 models, we access them through the official API at \url{https://openai.com/api/}, using the \textit{Text completion} endpoint. 
The considered model identifiers along with their sizes are:
\begin{itemize}
\footnotesize
    \item \texttt{text-ada-001}: 350M
    \item \texttt{text-babbage-001}: 1.3B
    \item \texttt{text-curie-001}: 6.7B
    \item \texttt{text-davinci-001}: 175B
    \item \texttt{text-davinci-003}: 175B

\end{itemize}

\begin{figure}[!t]
    \centering
    \includegraphics[width=\columnwidth]{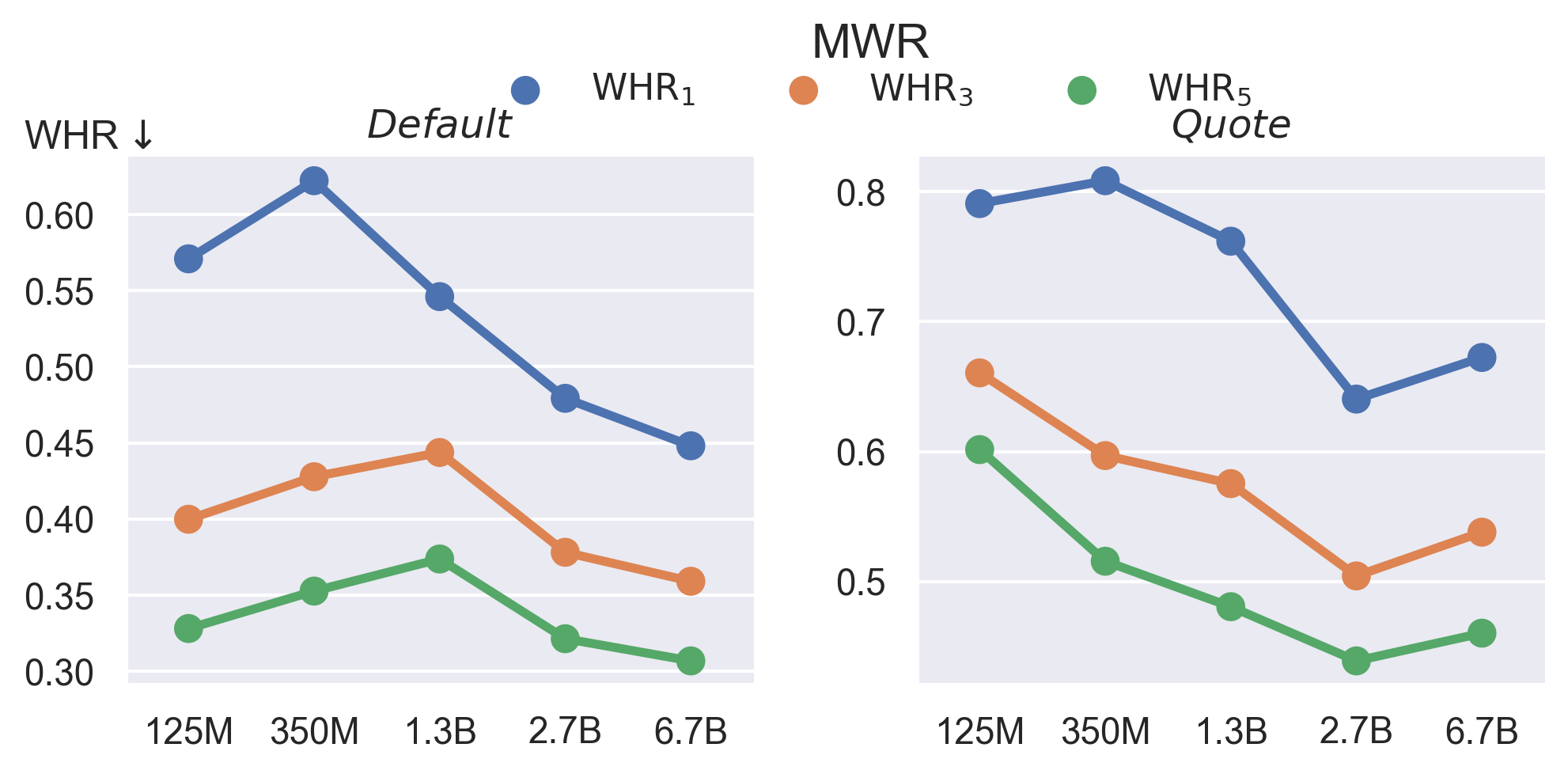} 

    \caption{Zero-shot performance of OPT on MWR using different prompts}%
    \label{fig:opt-mwr}
\end{figure}

\begin{figure}[!t]
    \centering
    \includegraphics[width=\columnwidth]{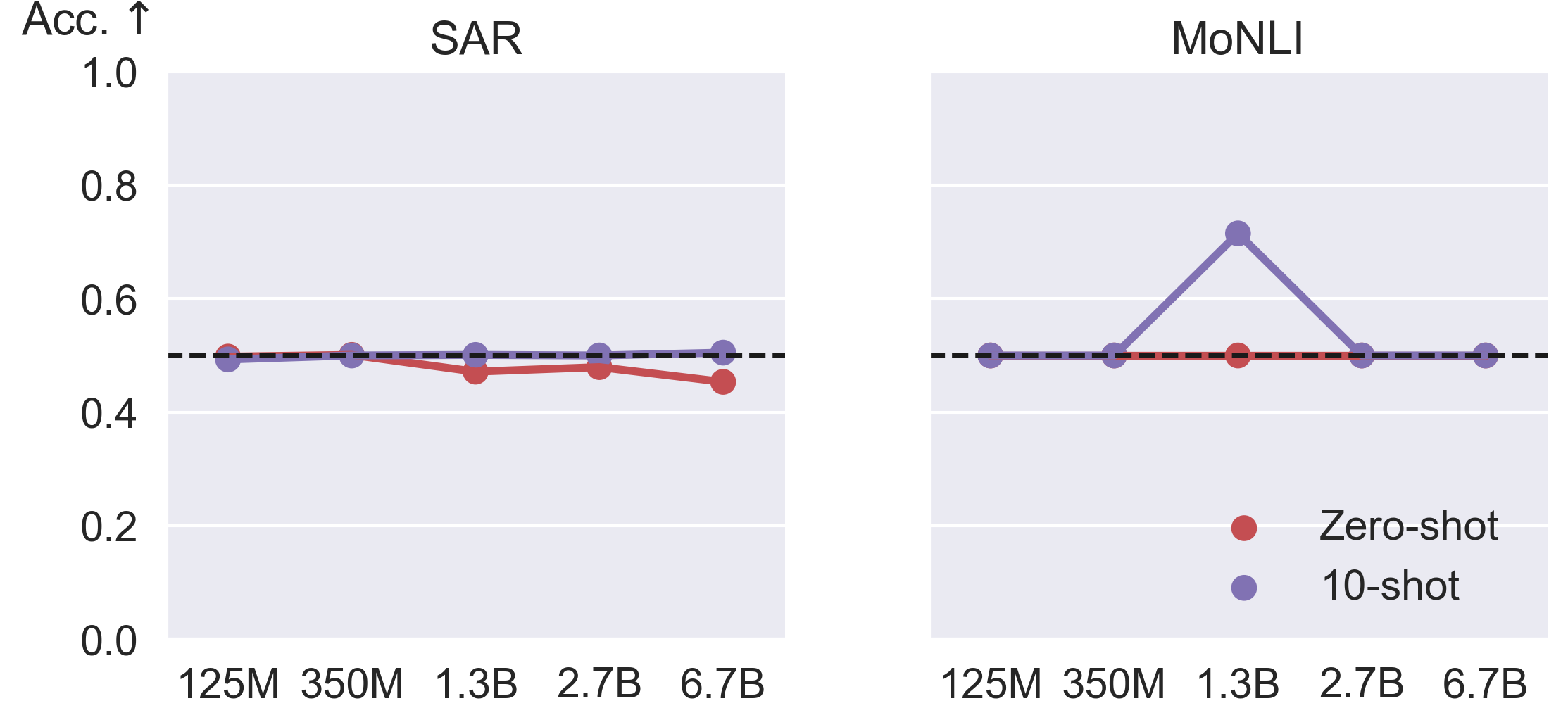} 

    \caption{Zero-shot performance of OPT on SAR using different prompts}%
    \label{fig:opt-fewshot}
\end{figure}

\section{OPT results}
\label{app:opt}

\begin{figure*}[!t]
    \centering
    \includegraphics[width=\linewidth]{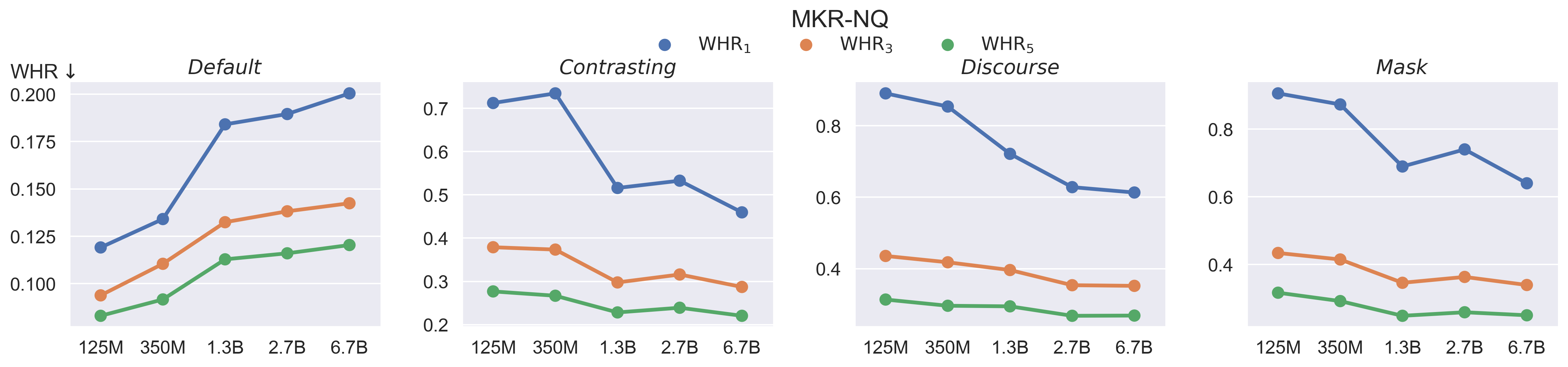} 

    \caption{Zero-shot performance of OPT on MKR using different prompts}%
    \label{fig:opt-mkr}
\end{figure*}

For MWR, although we observe improvements with increasing model sizes, the WHR scores are much higher than those of GPT-neo, showing that OPT is worse at predicting antonyms and synonyms of words.
The gap in performance may lie in differences in training data between the two types of models.

\begin{figure*}[!t]
    \centering
    \includegraphics[width=\linewidth]{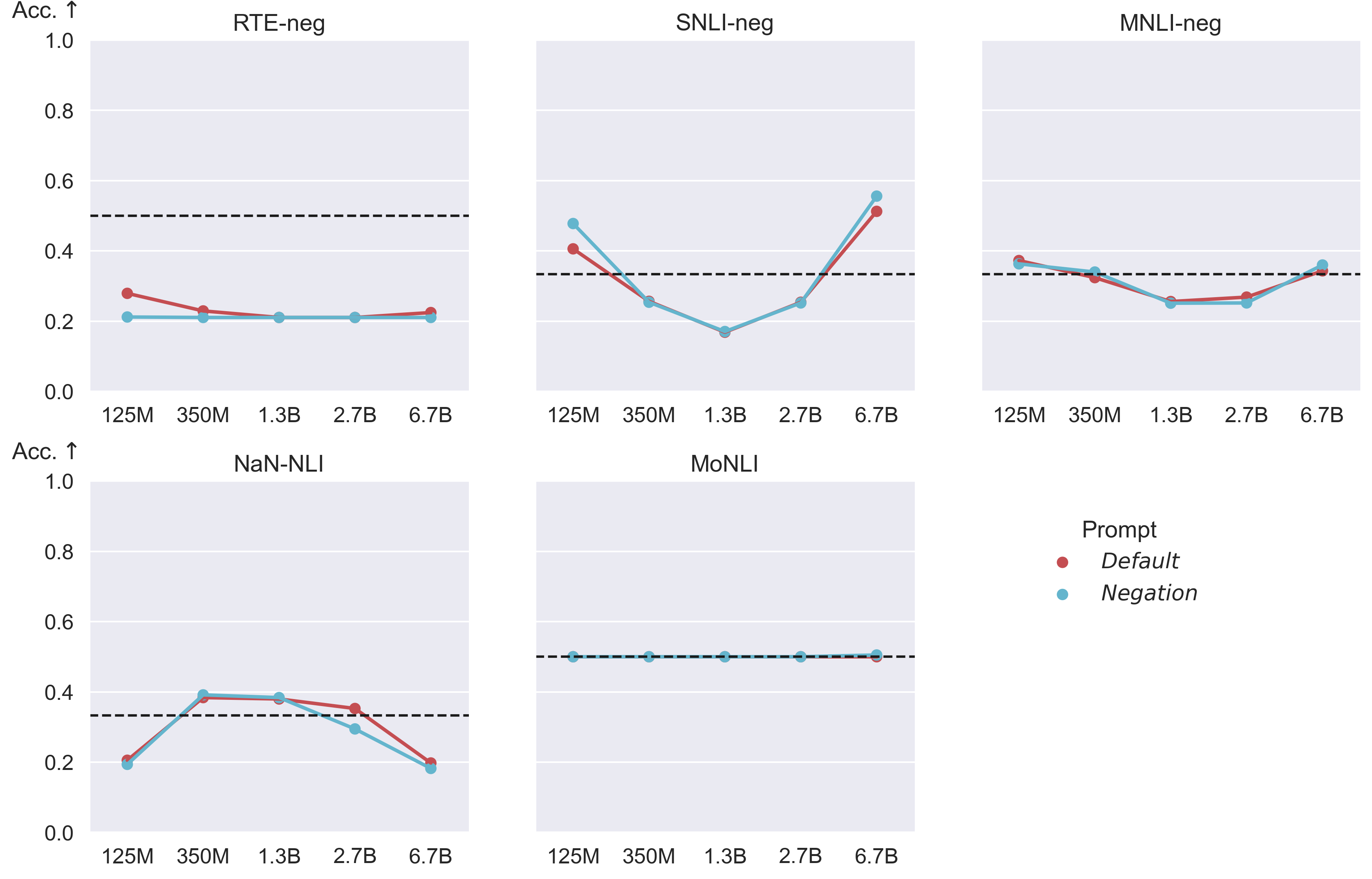} 

    \caption{Zero-shot performance of OPT on NLI tasks using different prompts}%
    \label{fig:opt-nli}
\end{figure*}


\section{Model outputs}
\label{app:outputs}

\begin{table*}[!t]
\footnotesize
\centering
    \begin{tabular}{p{9cm} c c}
        Prompt & Model & Output  \\
\midrule
      \ex{Paracetamol isn't a kind of} &  GPT-neo-125M & \ex{muscle}  \\
              & GPT-J-6B & \hl{painkiller} \\
              & OPT-125M & pain \\
              & OPT-6.7B & \hl{medicine}  \\
              & GPT-3 & \hl{medication}  \\
              & InstructGPT & \hl{NSAID} \\
\midrule
\ex{Entrance is an antonym of}  & GPT-neo-125M &  \ex{interest} \\
              & GPT-J-6B & \hl{entrance}  \\
              & OPT-125M & \hl{entrance}  \\
              & OPT-6.7B & exit  \\
              & GPT-3 & departure \\
              & InstructGPT & \hl{entrance} \\
\midrule
\multirow{6}{9cm}{\ex{Choose the correct answer: flimsy and sturdy are synonyms or antonyms?}}   & GPT-neo-125M & \hl{Synonyms} \\
              & GPT-J-6B & \hl{Synonyms}  \\
              & OPT-125M & Antonyms \\
              & OPT-6.7B & \hl{Synonyms} \\
              & GPT-3 & Antonyms \\
              & InstructGPT & Antonyms \\
\midrule

\multirow{6}{9cm}{\ex{I can not think of a few reasons for the allergy to substance. Question: There are not reasons why there's an allergy. True, False, or Neither? Answer:}} & GPT-neo-125M & \hl{True}  \\
              & GPT-J-6B & \hl{True}   \\
              & OPT-125M & \hl{True} \\
              & OPT-6.7B & Neither \\
              & GPT-3 & \hl{False}  \\
              & InstructGPT & Neither  \\
              \midrule
\multirow{6}{9cm}{\ex{The man does not own a dog. Question: the man does not own a mammal. True or Not true? Answer:}} & GPT-neo-125M & \hl{True}   \\
              & GPT-J-6B & \hl{True}   \\
              & OPT-125M & \hl{True} \\
              & OPT-6.7B & \hl{True} \\
              & GPT-3 & \hl{True} \\
              & InstructGPT & Not True  \\

    \end{tabular}
    \caption{Example outputs of models. Wrong answers are \hl{highlighted}}
    \label{tab:mkrnq}
\end{table*}





\end{document}